\documentclass[11pt,a4paper]{article}
\usepackage{times,latexsym}
\usepackage{url}
\usepackage[T1]{fontenc}

\usepackage{acl}
\usepackage{xspace,mfirstuc,tabulary}

\newif\iftaclinstructions
\taclinstructionsfalse 
\iftaclinstructions

\newcommand{\instr}
\fi

\newcommand{\data}{\texttt{PrimaryData}}
\newcommand{\generated}{\texttt{GeneratedText}}

\usepackage{tcolorbox}
\usepackage{tabularx}
\usepackage{multirow}

\usepackage{times}
\usepackage{latexsym}

\usepackage{comment}

\usepackage[T1]{fontenc}

\usepackage[utf8]{inputenc}

\usepackage{microtype}

\usepackage{inconsolata}
\usepackage{fontawesome}
\usepackage{booktabs}

\usepackage{marvosym}
\usepackage[disable]{todonotes}

\usepackage[]{titlesec}
\titlespacing*{\paragraph}{0pt}{1ex}{1ex}

\makeatletter
\newcommand*\iftodonotes{\if@todonotes@disabled\expandafter\@secondoftwo\else\expandafter\@firstoftwo\fi}  
\makeatother

\usepackage{hyperref}
\usepackage{xcolor}

\definecolor{darkblue}{rgb}{0, 0, 0.5}
\hypersetup{colorlinks=true,citecolor=darkblue, linkcolor=darkblue, urlcolor=darkblue}

\author{Harvey Lederman\\Department of Philosophy\\The University of Texas at Austin\\ \texttt{harvey.lederman@utexas.edu} \And  Kyle Mahowald \\ Department of Linguistics \\The University of Texas at Austin \\\texttt{kyle@utexas.edu}}

\title{Are Language Models More Like Libraries or Like Librarians? Bibliotechnism, the Novel Reference Problem, and the Attitudes of LLMs}

\begin{document}

\maketitle

\begin{abstract}
Are LLMs cultural technologies like photocopiers or printing presses, which transmit information but cannot create new content?
A challenge for this idea, which we call \emph{bibliotechnism}, is that LLMs generate novel text.
We begin with a defense of bibliotechnism, 
showing how even novel text may inherit its meaning from original human-generated text.
We then argue that bibliotechnism faces an independent 
challenge from examples in which LLMs generate \textit{novel reference}, using new names to refer to new entities.
Such examples could be explained if LLMs were not cultural technologies but had beliefs, desires, and intentions.
According to \emph{interpretationism} in the philosophy of mind, a system has such attitudes if and only if its behavior is well explained by the hypothesis that it does.
Interpretationists may hold that LLMs have attitudes, and thus have a simple solution to the novel reference problem.
We emphasize, however, that interpretationism is compatible with very simple creatures having attitudes and differs sharply from views that presuppose these attitudes require consciousness, sentience, or intelligence (topics about which we make no claims).
\end{abstract}

\section{Introduction}

Do modern LLMs have beliefs, desires, and intentions? Over the last few years, this question has been much discussed \citep[e.g.,][]{hase-etal-2023-methods,levinstein2023lie,shanahan2023role,mahowald2024dissociating,milliere2024philosophical,yildirim2024task}.
The hypothesis that LLMs do have these states is attractive in part because it offers a natural tool for explaining their behavior. It is standard to explain the complex behavior of humans and non-human animals in terms of what they think (believe), what they want (desire), and what they intend.
 If modern LLMs have beliefs, desires, and intentions, we can use the same tools to explain their behavior as well.

A challenge for those who deny that current LLMs have beliefs, desires, and intentions, is to provide an alternative, equally powerful, explanation of their behavior. 
Alison Gopnik and her coauthors have articulated a striking idea in this direction \citep{gopnik2022b,gopnik2022a,yiu2023transmission}. In Gopnik's view, LLMs are a
``cultural technology'', like a library or a printing press. Along these lines, the writer Ted Chiang compares prompting an LLM to ``searching over a library's contents for passages that are close to the prompt, and sampling
from what follows'' \citep{chiang2023chatgpt}.
Cosma Shalizi has dubbed this general idea ``Gopnikism'' \citep{shalizi}.
Since we will develop the position in our own way, we have given our version a new name:    \textit{bibliotechnism},  a combination of the Greek word for ``book'' and the Greek word for ``skill''. The defining commitments of bibliotechnism are, first, that LLMs are cultural technologies; and, second that LLMs do not have beliefs, desires, and intentions. The second claim especially will be key here, though it requires expanding Gopnik's position.

Can this view provide an explanation of the behavior of LLMs, which is sufficiently powerful to compete with the hypothesis that they have beliefs, desires, and intentions? 
We argue that if bibliotechnism is true, then if LLM-produced text is meaningful, its meaningfulness must in an important sense depend on the fact that text in the LLM's training data is meaningful.
If LLMs generated meaningful text, but its meaning were not of this ``derivative'' kind, then there would be an important sense in which, contrary to bibliotechnism, LLMs do not simply transmit existing cultural knowledge.
 
In normal cases, text produced by photocopiers and printing presses has only derivative meaning: it is simply a reproduction of human-generated input. 
But LLMs often produce entirely novel text, which is still apparently meaningful.
This seems to present a serious challenge for bibliotechnism: if LLM-generated text is meaningful only when it piggybacks on human-generated originals, how could such novel text be meaningful?

In Part I, we defend bibliotechnism against this challenge. Using n-grams as a toy model, and working up to more complex modern LLMs, we show how even novel text produced by LLMs may nevertheless derive its meaningfulness from the meaningfulness of the LLM's inputs.
We see this as an important step forward for bibliotechnism.

But in Part II, we present a further challenge for this view.
Modern LLMs are not just capable of producing novel text using old words, they also can produce newly invented names which apparently refer to newly created objects.
These new names cannot derive their reference from original text, since by assumption the name is not used in the data to refer to the relevant object.
We argue that this \textit{novel reference problem} poses a serious challenge for bibliotechnism.

LLM behavior in the novel reference problem would be easily explained if LLMs had beliefs, desires, and intentions. 
According to \emph{interpretationism} in the philosophy of mind and cognitive science, a system has beliefs, desires, and intentions if and only if its behavior is well explained by the hypothesis that it has such states  \citep[e.g.,][]{dennett1971intentional,dennett1989intentional,davidson1973radical,davidson1986coherence}.
According to this thesis, there is no difference between the question of whether a system's behavior is well-explained \emph{on the hypothesis} that it has certain attitudes, and the question of whether it \emph{actually has} those attitudes.
We argue that, if this philosophical position is true, then there is evidence that LLMs have beliefs, desires, and intentions, and the novel reference problem can be easily solved.
We emphasize, however, that interpretationism differs in important ways from common lay views about the ontology of attitudes. It does not require that a system have particular internal correlates (say, a particular neuron) of putative beliefs or intentions; nor does it require that a system be intelligent or conscious in order to have attitudes. Instead, a system may count as having beliefs or intentions because of patterns visible in its global behavior. Thus, for interpretationists, attributing beliefs, desires, and intentions to LLMs does \emph{not} require that they are sentient or conscious \citep[see][for discussion of these questions]{chalmers2023could,butlin2023consciousness,goldstein2024zombies}.

\section{Prior Work and Background}

\paragraph{Prior Work}

A prominent line of argument has suggested that LLMs cannot produce reference without being ``grounded'' \citep[e.g., ][]{bisk-etal-2020-experience,lake2023word}.
\citet{bender-koller-2020-climbing} have influentially argued that LLMs cannot understand language in part because they cannot have perceptual contact with objects to which speakers intend to refer, and so cannot learn those speakers' communicative intents. A natural inference from their discussion is that, owing to LLMs' inability to understand language, they can also not produce meaningful text \citep[cf.][]{titus2024chat}.

\citet{piantadosi2022meaning} propose an alternative account of meaning in which meanings are constituted by the relationship among concepts in a particular conceptual space.
Since LLMs clearly represent rich inferential relationships as well as relations of semantic similarity, in their view LLMs can produce meaningful words even without perceptual exposure to their referents.

\citet{mandelkern2023language} connect the debate between \citet{bender-koller-2020-climbing} and \citet{piantadosi2022meaning} to semantic externalism, a view of meaning which has been dominant in the philosophy of language since the 1980s. On a standard view, \citep{putnam1975meaning,kripke1980naming,burge1986individualism}, people can refer to Shakespeare without having been in direct causal contact with Shakespeare, in part by belonging to a community whose overall use of this word stands in an appropriate causal relationship to the poet. 
Mandelkern and Linzen accordingly suggest that whether LLMs can produce text that refers to Shakespeare depends on whether LLMs ``belong to our speech community''.

\citet{cappelen2021making} earlier provided a systematic application of semantic externalism to model-generated text. They argue that such text can be meaningful, but that in order to accommodate the full range of model behavior, we must revise and extend standard theories of meaning. In response, they propose a novel  \emph{interpreter-focused} ``metametasemantics'', according to which the meaningfulness of novel generated text depends on what someone could come to know by reading it. 

\citet{nefdt2023contextualist} similarly sees LLMs as inspiring revision to the theory of meaning, suggesting the meaningfulness of model-generated text requires a form of radical contextualism.

The present paper goes beyond these earlier works by considering the relationship between the meaningfulness of LLM-generated text and the viability of bibliotechnism.
Moreover, earlier theoretical work does not clearly vindicate the claim that n-gram models produce meaningful text, whereas we will argue that they do.
Moreover, we do so without revision to standard theories of meaning.
We build on our account to defend bibliotechnism against an important challenge, demonstrating that the view can accommodate the meaningfulness of even entirely novel text generated by LLMs.
We then introduce the novel reference problem as a new challenge for bibliotechnism.

\paragraph{Background}

Modern generative language models are trained on \data{}, typically on a word prediction task and sometimes with additional training to bring about particular behaviors  \citep[e.g., Reinforcement Learning from Human Feedback (RLHF) whereby models receive explicit feedback to align with human preferences][]{ouyang2022training}. 
At generation time, they are prompted and then sampled to probabilistically produce \generated.
We will take it as uncontroversial that words in \data{} are meaningful and refer to things in the world.
For instance, if a human-authored biography of Shakespeare is included in \data{} and includes the line ``Shakespeare was born in 1564.'', the relevant token of ``Shakespeare'' is meaningful and refers to the poet.

What will matter for our purposes below is that the models are (a) trained on largely naturalistic human data to generate text, (b) do not simply verbatim reproduce their training data, and (c) produce text which human users can typically understand.
Today's LLMs 
have these properties: they are trained on human data, generate at least some novel text 
\citep{mccoy-etal-2023-much}, and produce fluent text that human users engage with easily.

We highlight three points about philosophical terminology.
Philosophers often distinguish between ``reference'' and ``meaning''. As we will use the terms here, any expression that refers has a meaning, although many meaningful expressions do not refer.
For simplicity, the only words that we will assume can \emph{refer} are proper names. 
Those unused to this distinction may treat ``meaning'' and ``reference'' as synonymous throughout; nothing essential will turn on the difference between them.

Second, there is a difference between the word ``Shakespeare'' and particular \emph{tokens} of this word. If the word ``Shakespeare'' is written on a blackboard five times, there are five tokens of this one word on the blackboard. Words, complex expressions, and strings more generally are \emph{types} which can occur in multiple ``tokens'', whether in written inscriptions or spoken utterances. We assume that both tokens of words and of longer strings can be meaningful.

Third, and finally, we distinguish between ``referring'' and ``meaning'' that is done by agents (``Marlowe was referring to Queen Elizabeth.''), and referring and meaning that is done by particular word-tokens. Since our goal is to explore a view on which LLMs are not agents, we will be investigating the question of whether such non-agents can produce tokens which refer and are meaningful. We will not be assuming that LLMs themselves can refer or mean \emph{as agents}.

\section{Part I: How to be a Bibliotechnist}

\subsection{Bibliotechnism and Derivative Meaning}

Bibliotechnists take cultural technologies, like books and libraries, to be tools for the transmission and dissemination of information, allowing the accumulation of knowledge over large stretches of space and time. These technologies are, crucially, not themselves responsible for new ideas or information.
They simply transmit information which already existed.

Bibliotechnists take LLMs to be cultural technologies, and, accordingly, hold that if LLMs generate meaningful text it must be what we will call \emph{derivatively} meaningful. 
When an author writes ``Shakespeare'', their inscription of the word (the word-token) is meaningful and refers to the poet.
As a result of this case of reference, the token of ``Shakespeare'' on the 13th page of the 423rd copy of the 3rd printing of this biography \emph{also} refers to the poet.
The same holds also for tokens of ``Shakespeare'' on photocopies of this page of this edition of the book, even if the photocopies are produced by accident. 
A similar thesis applies also to the meaning of complex expressions (involving more than one word) like ``Shakespeare was born in 1564''. 

We will say that tokens created immediately by an agent are instances of \emph{basic} meaning and reference, while other tokens are instances of \emph{derivative} meaning and reference. We stipulate that it is only agents---that is, entities that have beliefs, desires, and intentions---that produce tokens which refer or are meaningful \textit{basically}.
Beyond this stipulation, the distinction between basic and derivative reference and meaning is rough, but we will only deal with clear examples of each category in what follows.

According to bibliotechnism, LLMs do not have beliefs, desires, or intentions. So, according to bibliotechnism, LLMs can only produce tokens which refer or are meaningful derivatively. 
Is this consequence of the position correct?
We first argue that unigram models can produce \emph{words} which refer and are meaningful derivatively, but that they cannot produce \emph{complex expressions} which are derivatively meaningful.
We then explain how, going beyond unigrams, LLMs can produce complex expressions---and indeed long stretches of entirely novel text---which are derivatively meaningful.

\subsection{Causal History and Derivative Meaning}

In this section, we argue that derivative reference and meaning can be achieved by an appropriate causal connection between \data{} and \generated{}, and show how this vindicates the idea that n-grams can produce derivatively meaningful word-tokens. 

Since the 1970s, philosophers have developed the idea that causal connection can play a key role in facilitating reference and meaning, and that our ability to refer to, say, Shakespeare is partly explained by there being an extended causal chain from current humans, through their teachers, their teachers' teachers, and so on, all the way back to the poet \citep{geach1969perils,donnellan1970proper,evans1973causal,kripke1980naming}.

This relationship between current human uses and an original ``baptism'' is conceptually distinct from the relationship between derivatively meaningful tokens produced by a photocopier or printing press and \emph{basically meaningful} tokens produced by a human.
But in both cases it seems that causal connection is important. In particular, we suggest that derivative reference and meaning also depend on an appropriate causal chain tracing from a new token back to a token which is meaningful basically.
It is because the token of ``Shakespeare'' on the 13th page of the 423rd copy of the 3rd printing of the biography is appropriately causally connected to the original token created by the author, that it refers to the poet.
The same also goes for tokens of ``Shakespeare'' on photocopies of this page: these new tokens can refer because they are appropriately causally connected to a basically meaningful token. 

This ``appropriate'' causal connection does not require human supervision.
If a page from a book flies into a malfunctioning photocopier, making copies by accident, the tokens on the resulting page would still refer to the poet.
If whole sentences are copied by the machine, these sentences would also be derivatively meaningful, because of their causal connection to an original token.

This observation already shows that unsmoothed large-n n-gram models---which straightforwardly sample from a distribution conditional on the previous $n-1$ words---can produce meaningful tokens.
For sufficiently large $n$ (e.g. 1000), such models simply copy tokens from \data{}.
So, like a photocopier, they produce meaningful tokens.

Matters are less straightforward for unigram models, which sample from a distribution of single words.
We can think of such models as implemented by taking all of their \data{}, choosing word-tokens from the \data{} at random, and then copying the chosen token.
The tokens of individual words the n-gram then produces are again just like those of a copier (or of the large-n model): they have a direct causal connection to the original tokens. 
As a result, if the model produces a token of ``Shakespeare'', this token will be meaningful (and refer) derivatively, piggybacking on the meaning and reference of a token of this word found in \data{}.

In this case, however, there is a new phenomenon, not exhibited in the case of the photocopier or large-n n-gram. 
Each of the tokens of the individual words produced by the unigram will be meaningful, but it does not seem that tokens of complex expressions containing these tokens will be.
The vast majority of the time, the token the model produces will be a token of a gibberish string, and uncontroversially meaningless.
At low odds the unigram model will produce a token of a  ``reasonable'' string like ``Shakespeare was born in 1564''.
With even lower odds, a token of a reasonable string will be produced by copying an original token of that string.
But even in these latter cases, the fact that a token of a string which could be meaningful is produced is a fluke, a complete accident.
As a result, we judge that, even when the model produces a token of a string that would be meaningful if produced by a human in a normal way, the token the model produces is not meaningful.

The reason is that the generated text is not appropriately causally connected to the \data{} to inherit the ``glue'' that binds the words in the complex expression together.
 The tokens are like sand 
 blown into the shape of a sentence by the wind.

\subsection{Derivatively Meaningful Expressions}

Thus far, we have seen how tokens of complex expressions can be derivatively meaningful if they are copied from \data{}. But modern LLMs routinely produce tokens of strings which do not appear in \data{}.
Can bibliotechnism accommodate the meaningfulness of such novel text?

We will argue that it can, by arguing that tokens of complex expressions can be derivatively meaningful because of two distinct causal pathways involved in their production: a first (discussed above), which connects new tokens of individual words to originals in the data; and a second (new to this section), which guarantees that outputs possess higher-level features of expressions in \data{} to serve as a kind of ``glue'' binding word-tokens into meaningful tokens of longer strings.

To see the idea, consider a rudimentary model, which, when fed a token of a sentence, finds any name-tokens in the sentence (searching with a database) and then replaces each of these names uniformly with a name drawn at random from the distribution of all names in its \data{}. 
This model plausibly produces not just tokens of individual words which are derivatively meaningful, but also can produce tokens of never-before-considered strings which would, as a whole, be derivatively meaningful. 
For instance, if we gave this model our Shakespeare sentence, and it produced ``Plato was born in 1564'', this token would be false, but meaningful, even if it is a token of a string never before contemplated by a human.
Here, the causal history of the context and the causal history of the individual name-token are different, but the whole token expression would still be derivatively meaningful.
Plausibly, this is because the operation as a whole is \emph{causally sensitive} to a structural, high-level feature of its input, and in particular, the fact that it reliably produces an output sentence that preserves the grammatical structure of its input sentence.
A rough test for causal sensitivity in this sense (though not a necessary or sufficient condition) appeals to counterfactual sensitivity: (i) if the model were given \data{} exhibiting property $P$, would it reliably produce outputs which exhibit property $P$?; and (ii) if the model were given \data{} which does not exhibit property $P$, would it reliably produce outputs which do not exhibit property $P$?
The rudimentary model just described passes this test, and is in any case intuitively causally sensitive to the structure of its input.
It is in part owing to this causal sensitivity that the model can produce tokens of novel sentences which are derivatively meaningful, since the output tokens inherit from their input not only the meaningfulness of their constituents, but also their form.

This discussion provides one example where novel text can nevertheless be derivatively meaningful. 
It also suggests that, in general, if a model is causally sensitive to relevant high-level features of its \data{}, in such a way as to transmit those features to its \generated{}, it may produce even entirely novel text that is nonetheless derivatively meaningful.
The question then becomes: are modern LLMs sensitive to an appropriate high-level feature?

A first proposal might focus on \emph{grammaticality}. 
One might suppose that it would suffice for an LLM to produce meaningful output, if it is causally sensitive to the grammaticality of its \data{} in such a way that it reliably produces grammatical \generated{}.
But this proposal is not correct in general, because not all grammatical expressions are meaningful.
In fact, even the simple operation of replacing  constituents with others of the same syntactic category can lead to expressions that, while grammatical, are meaningless, e.g.: ``You apply the toy and serve fighter hair into the blackmail'' \citep{gulordava-etal-2018-colorless}.
So, a process that reliably produces grammatical sentences with derivatively meaningful sub-expressions could still fail to produce derivatively meaningful sentences.
In short, if LLM-produced text is meaningful, it cannot be only in virtue of causal sensitivity to the grammaticality of its input.

We suggest that a more promising high-level property is what we call \emph{intelligibility}. In our technical sense, a token of a string is intelligible in a language if and only if it is possible that someone understand a token of the string in line with the conventions of the language. (From now on ``intelligibility'' will be ``intelligibility in English''.) Intelligibility in this sense does not vary from person to person: it depends only on whether it is possible that \emph{someone} understand the sentence, not on whether one particular person does. And a token which is intelligible in this sense need not be meaningful: a token formed by wind in the sand may be intelligible (if it is a token of a string, some tokens of which are understood), even if it is not meaningful itself.

 We suggest that, if LLMs are causally sensitive to the intelligibility of their input, in such a way as to produce intelligible outputs when given intelligible inputs, then tokens of intelligible complex expressions they generate will be derivatively meaningful, with individual words inheriting the meanings of the original tokens from which they were ``copied'', and whole expressions inheriting the ``glue'' of intelligibility from the \data{}.
The rough counterfactual test for causal sensitivity discussed above suggests that modern LLMs are in fact causally sensitive to the intelligibility of their \data{}. First, these models overwhelmingly produce text which is intelligible.
Second, if LLMs were trained on gibberish, they would output gibberish. 
Given that modern LLMs are causally sensitive to intelligibility in their input, there is a clear story according to which modern LLMs do not just produce derivatively meaningful single words like unigrams, but in fact can produce derivatively meaningful complex expressions.

We will continue to develop bibliotechnism using the property of intelligibility. But there are many other candidate notions that one might use instead: semantic well-formedness, discourse coherence, or even felicity. We ourselves think none of these quite does the needed work.  But nothing essential in what follows will turn on the exact property chosen, and a reader who prefers these other properties may use them in place of intelligibility.

Intelligibility in our sense does not require truth, and causal sensitivity to intelligibility does not require reliable production of the truth. 
Tokens of false sentences like ``Shakespeare was born in 2023'' are intelligible.
Even the best LLMs at the time of writing are known to produce text that confabulates or fabricates information. 
But getting a fact wrong (e.g., saying Shakespeare was born in 2023 instead of 1564) is importantly different than producing unintelligible text.
If an LLM reliably responded to queries about Shakespeare's birth with gibberish, this would at least be some evidence that it is not in fact causally sensitive to the intelligibility of its data in such a way as to generate derivatively meaningful complex expressions.
But producing false, intelligible text is entirely compatible with such causal sensitivity.

It is instructive to compare LLM-generated novel text to text generated by bigram or trigram models.
Unlike unigrams, bigram and trigram models fairly reliably copy short complex phrases from their input. Indeed, such models might be statistically likely to combine expressions in ways that might seem meaningful as a whole, because they are causally sensitive to certain features of the patterns of combination of these words in the data.
For instance, a bigram model that outputs ``Shakespeare wrote plays'' does so in part because it is sensitive to the fact that ``wrote'' is a likely continuation for ``Shakespeare'' and that ``plays'' is a likely continuation for ``wrote''. 
But the only causal connection between its production of ``Shakespeare'' and its production of ``plays'' is mediated by the verb ``wrote''.
Given this fact, we judge that the causal story about its production of this sentence does not preserve an appropriate connection between all parts of the sentence.
Accordingly, even when the model produces tokens of sentence-length strings that are grammatical, and even when individual phrases may be judged meaningful, it seems that, as with unigram models, longer sentences should probably not be understood as meaningful.
These sentences lack the straightforward ``copy property'' of higher-n n-gram models but also are not produced in a way that is causally sensitive to relevant structural features (and in particular intelligibility), as text generated by modern LLMs seems to be.

We conclude that LLMs can produce novel text which is nevertheless derivatively meaningful, because they copy individual tokens from their \data{}, and assemble them in ways that are causally sensitive to the high-level feature of intelligibility in their \data{}.

Before closing this discussion, we offer one key clarification about the basis of our judgment that unigrams do not produce derivatively meaningful complex expressions.
The basis for this judgment is \emph{not} the fact that n-grams are only trained on individual words.
It is instead because n-grams are not causally sensitive to relevant high-level features of their \data{}.
That is, we are not interested in a narrow form of ``input-sensitivity'', but instead in a broader notion of causal sensitivity, partly captured by the test of counterfactual sensitivity.

This contrast can be illustrated by considering again a photocopier. The fact that a photocopier responds to (say) features of the ink used to write original letters is irrelevant to the question of whether the tokens it produces are meaningful.
As long as its underlying low-level mechanism leads to causal sensitivity to the right high-level features---as evidenced in this case by the fact that it reliably produces tokens of words when it is fed words, and reliably does not produce words when it is not---the tokens it produces will be derivatively meaningful.

The same point can be made in connection to an n-gram trained not on word frequency but on letter frequency.
In fact, a unigram model trained on letters (as opposed to words) with the same \data{} as the models above, would in its trained form do nothing more than spit out letters randomly in proportion to their frequency in \data.
But if a letter-tokenized n-gram model somehow \emph{were} sufficiently reliable in producing real words (as a 10-gram model trained over letters might be), that would be evidence that it was causally sensitive to the fact that letters in its \data{} formed words, and that it was producing derivatively meaningful tokens of these words. 
In short, an n-gram trained on letters may fail to produce referring tokens not because it is trained on letters, but because that training mechanism (as a matter of fact) is not causally sensitive to the right high-level features of its \data{}. 

This concludes our response to the first challenge for bibliotechnism, that LLMs can produce novel text which is apparently meaningful. 
We see this as an important step forward for the tenability of bibliotechnism. But we are not yet convinced that the view is correct, and we now turn, in the rest of the paper, to a new and different kind of challenge to it: the novel reference problem.

\section{Part II: A Problem for Bibliotechnism}

\subsection{The Novel Reference Problem and Replies}

We will illustrate the problem of novel reference with two examples.
The first example involves cases where LLMs would produce tokens of names they have never seen before, intuitively in such a way that they refer to previously referred-to objects.
 In this task, we ask an LLM to choose any real historical figure it likes, and then come up with a new name and tell us facts about this historical figure 
(see Appendix \ref{prompt} for details).
ChatGPT (GPT-4) completed this task using ``Marion Starlight'', in text describing a figure ``born in the 18th century'', who ``authored a famous pamphlet that criticized the French monarchy'', ``played a critical role in the French Revolution'', ``became increasingly paranoid and was involved in the Committee of Public Safety, which oversaw the Reign of Terror'', and ``was arrested and executed during the Thermidorian Reaction.''
The tokens of ``Marion Starlight'' in this text plausibly refer to the historical figure Robespierre, 
even though, as a stipulation about what counts as success in the task, there are no tokens of ``Marion Starlight'' in the data which refer to Robespierre.
 If the task was performed successfully, the tokens of this name could not refer to Robespierre in virtue of basic reference exhibited by tokens of this name in the \data{}.

A second, more powerful example, sharpens the problem. 
Here, an LLM is tasked to produce an ASCII picture which it has never seen before, to give elements of the picture names, and then to describe the picture using those names. 
If the LLM succeeds, then it is even clearer than in the previous example that the reference of relevant expressions could not be due to reference of the relevant name in the \data{}, since the object did not exist in this form until the LLM created it. 
Insofar as LLMs have empirically been shown to be able to generate, designate, and manipulate elements of code-generated pictures \citep{bubeck} and also to refer meaningfully to novel orientations of elements in visual and color spaces \citep{patel}, the ability to complete this task seems solidly within their capabilities.

If LLMs can perform successfully as expected in such examples, then the meaningfulness of the tokens they produce cannot be straightforwardly accommodated by the account of derivative meaning and reference that we have given, since by assumption no tokens of the relevant (novel) names in \data{} refer to the relevant entities.

But if bibliotechnism is correct, and these tokens are meaningful, they must be cases of derivative meaning, so there must be some way in which they ``piggyback'' on basic human meaning and reference.
Other than \data, which we already ruled out, there are four salient ways that human attitudes might enter to guarantee the meaningfulness of the new inscriptions. 
We consider responses to the novel reference problem based on these four possibilities below.

\paragraph{Human Feedback in RLHF} 

\citet{mollo2023vector} claim that the RLHF step, in which models receive human feedback that pushes them towards desirable behaviors, is critical for the meaningfulness of LLM \generated{}.
Their idea is that human intentions may ground LLM reference by  ``aligning'' the LLM with human goals \citep{bai2022constitutional}.
But while RLHF influences model capabilities, even models without RLHF produce apparently meaningful text.
Since the un-RLHF-ed models plausibly produce this text in a manner which is causally sensitive to the intelligibility of their \data{}, our earlier account predicts that they too can produce meaningful text.
An account which considers RLHF-ed models to be radically different in their basic referential abilities fails to deliver this verdict, and thus fails to accommodate the apparent meaningfulness of the text they produce.
Moreover, this limitation would plausibly also be present in an account of novel reference which depended centrally on RLHF, since we conjecture that models which have not undergone RLHF can perform the task sufficiently well that their output would have an equal claim to be meaningful as models which have.

\paragraph{Creators' Intentions}
A second point at which intentions might enter the pipeline is during the creation of the LLM.
A very precise thermometer may produce a token reporting a temperature that no one has ever thought about, and this token seems to ``refer'' to this temperature.
Its ability to do this seems to derive from the creator's general intention at the time of construction: that any indication using some numbers would refer to the relevant temperature.
By the same logic, one might say that an LLM's creators' intentions might be general enough to guarantee that the words it produces would be meaningful in the relevant language and perhaps to accommodate our cases of novel reference.
But even supposing this response were to offer an explanation of the capacity for novel reference in some LLMs, it is not sufficiently general to accommodate our judgments about all relevant models. 
LLMs can be created for different reasons: if the ``same'' LLM was created by Team A for the purpose of measuring sentence probabilities for use in a downstream application, and by Team B for use as a chatbot, it seems odd to conclude (as the present response requires) that only the second of these generates meaningful text in our cases.

\paragraph{Intentions in Generating the Prompt}
A third point at which human intentions might enter the production process is through the user. Perhaps in our particular prompts involving novel reference, the \textit{user} has an intention that whatever name the LLM produces (e.g, ``Marion Starlight'') should refer to the person best described by the surrounding text (or to the aspect of the diagram best described by this text).
On this view, the tokens produced by the LLM are only meaningful in virtue of the user's attitudes.
Whether or not this approach succeeds for actual LLMs today, it again does not make correct predictions in relevantly similar cases. Suppose that we initiate a process in which an LLM is provided with random prompts (perhaps prompts generated by a unigram model), with no intentions about the meaningfulness of any generated text. Suppose moreover that by chance a model is fed our prompt asking for a story featuring a new name for a historical figure. If the LLM produced the responses described above, it still seems to us that it would produce tokens which refer to Robespierre or to aspects of the relevant diagram. But this reference could not be due to the intentions of the creator of the prompt, since by assumption there is no user that has intentions.

\paragraph{Reader's Intentions}
A fourth and final place where human intentions might enter the picture is through the reader of the text (who might not be the creator of the prompt). In this vein, \citet[Ch. 4]{cappelen2021making} develop an interpreter-focused ``metametasemantics'' according to which tokens can count as meaningful in virtue of how readers would understand them. We consider this response the most promising option for bibliotechnists who take LLM-produced novel name-tokens to be meaningful.

But there is a serious problem for the response as it stands. A beachgoing bardolator might write a token of ``Shakespeare was born in 1564'' in the sand, in a manner that would appear visually exactly the same as a token of this string produced by the wind. Intuitively, the first token would be meaningful, while the second would not. But a simplistic interpreter-focused approach cannot respect this judgment; it predicts that one token is meaningful if and only if the other is.

The more sophisticated approach developed by Cappelen and Dever handles this case, by requiring that our attribution of meaning maximize the \emph{knowledge} of the interpreter. Since the wind's writing is too random to confer knowledge, they presumably would not say that the wind-produced token is meaningful. But their approach still cannot handle related examples. For instance, if a frog evolved to reliably croak only on nights prior to rain in a way that sounds like ``rain'' to English speakers, the frog's ``utterance'' would intuitively not have linguistic meaning. Still, an interpreter could use it to acquire knowledge of the next day's weather, so Cappelen and Dever's account would appear to predict that it has linguistic meaning. 

Receiver-focused theories offer a liberal account of when tokens are meaningful. This liberality allows them to accommodate novel reference by LLMs. But it also prevents them from distinguishing between tokens produced by humans, and those produced by the wind (or frogs). The challenge for such theories is to eliminate the prediction that wind-produced tokens are meaningful without eliminating the prediction that LLM-produced tokens are. We certainly have not shown that it is impossible to meet this challenge. But more work will be required to show that it can be met.

\subsection{Interpretationism and Attitudes}\label{attitudes}

How might the problem of novel reference relate to the broader question of whether LLMs have attitudes? In this section, we first introduce interpretationism, then consider its application to LLMs, and close by describing how it solves the novel reference problem.

\subsubsection{Interpretationism}

How do attitudes like belief, desire, and intention fit into explanations of human behavior?
Human behavior can be explained and predicted at the microphysical level by the laws of physics.
But this fact does not mean that beliefs, desires, and intentions are not \emph{also} useful in explaining and predicting behavior.
These descriptions are not as informative as complete microphysical descriptions.
But they are still good explanations: they are efficient tools for making high-level predictions about future behavior, without significant loss of accuracy.

According to \emph{interpretationism} in the philosophy of mind and cognitive science \citep[e.g.,][]{dennett1971intentional,davidson1973radical,davidson1986coherence,dennett1989intentional}, a system has beliefs, desires and intentions if and only if its behavior is well explained by the hypothesis that it has those attitudes.
Along these lines, \citet{mccarthy1979ascribing} writes that such attitudes are: ``legitimate when such an ascription expresses the same information about the machine that it expresses about a person'' and ``useful when the ascription helps us understand the structure of the machine, its past or future behavior, or how to repair or improve it.''

We will focus on three criteria for what makes behavior ``well explained'' by some hypothesis, in the technical sense used by interpretationists: (i) accuracy: that the hypothesis makes sufficiently accurate predictions; (ii) power: that the hypothesis makes predictions in a wide array of independently specifiable circumstances; and (iii) tractability: that its predictions are easy to derive. A hypothesis well explains some behavior if it does sufficiently well in these three categories taken together, where what counts as ``sufficiently well'' may depend on how it compares with alternative explanations. 

To illustrate: An explanation of an apple's fall in terms of its desire to fall to the ground is tractable, but not predictively powerful, since there is no independent specification of when it ``wants'' to fall to the ground (and, if there were, exploiting it would often lead to inaccurate predictions).
By comparison, an explanation of the apple's fall in terms of gravity is less tractable, but is much more predictively powerful and accurate. 
People are quite different from apples.
An explanation of a person's buying of a cup of coffee in terms of their desire for coffee, along with their belief that buying it is the best way to get it, is tractable and makes predictions in independently specifiable circumstances. We can predict that the person will not want more coffee after having a lot, that they may want coffee again the next day, and so on. 
Even though this theory is less accurate and less powerful than one which fully explains the human's behavior in terms of cellular biology or atomic physics, it makes up for these losses by enormous gains in tractability.
To predict whether a friend will want coffee, atomic physics is simply not a practical tool.
In light of these facts, interpretationists hold that the behavior of people but not the behavior of apples is well explained by the hypothesis that they have beliefs, desires, and intentions. So interpretationism correctly predicts that the former, but not the latter, actually have such attitudes.

Interpretationists hold that people can be mistaken about whether an explanation is in fact a good one. An ancient animist may have believed that the fall of an apple was well explained by its desire to fall. But interpretationists hold that this was not a good explanation, in spite of what the animist thought. 
Similarly, if ELIZA passed the Turing Test in an hour-long conversation with a person, the person might claim that the best explanation of their interlocutor's behavior was that it has beliefs, desires, and intentions.  But they would be wrong. 
An explanation of ELIZA's behavior in terms of its being a lookup table is more accurate and powerful (predicting various mistakes and failures) than the hypothesis that ELIZA has beliefs, desires, and intentions. 
Interpretationists will say that, notwithstanding what the person thought, ELIZA's behavior is not well-explained by the hypothesis that it has attitudes, and, thus, conclude that ELIZA does not have them.

 Most of human physiology and even some human behavior is not well-explained by our beliefs, desires, and intentions: shivering, digesting, speech errors, and invented memories. Each of these behaviors can be well explained in deeper, biological terms. But explanations using attitudes either don't apply (shivering, digesting), or give the wrong result (speech errors, invented memories). Still, these cases do not show that people do not have attitudes. As long as the relevant part of our behavior is sufficiently well explained by this hypothesis (allowing some inaccuracy, but not too much), interpretationists will still say that people do in fact have beliefs, desires, and intentions.

There is no universally accepted philosophical theory of the nature of belief, desire, and intention.
Certainly, not all philosophers accept interpretationism \citep[for a survey of alternatives see][]{sep-belief}. But the view has important attractions. Ancient people reasonably and correctly attributed beliefs, desires, and intentions to one another in spite of mistaken beliefs about human biology. 
 Plausibly, an alien species with entirely different internal architecture, but outwardly similar behavior, could have beliefs, desires, and intentions. Interpretationism makes sense of these intuitive data, by understanding attitudes in terms of patterns of behavior, not in terms of the details of internal architecture.

\subsubsection{Interpretationism and LLMs}

Interpretationism is of interest in the present context because some interpretationists will think that there is already a case to be made that LLMs have beliefs, desires, and intentions.
LLMs have shown (imperfect) success in: solving math word problems \citep{lewk}, keeping track of entities  \citep{kim-schuster-2023-entity}, and reasoning about social situations \citep{trott2023large}.
In these cases, an explanation in terms of their being next-word predictors may be accurate, but it is not particularly tractable. A far more tractable, but still mostly accurate and powerful explanation of these behaviors can be given on the assumption that LLMs have beliefs, desires, and intentions---even while recognizing that LLM performance is imperfect and/or un-humanlike in each case. So interpretationists may hold that there is an independent case that LLM behavior is ``well explained'' (in the technical sense) by the hypothesis that LLMs have attitudes, and thus, given their distinctive view of these attitudes, that LLMs do have them.

Of course, not all LLM behavior is well explained by the hypothesis that they have beliefs, desires, and intentions. 
Models are sometimes ``right for the wrong reasons'' \citep{mccoy-etal-2019-right} and might, say, perform a logical entailment task by detecting spurious statistical correlations in training.
They are often inconsistent in their beliefs \citep{hase-etal-2023-methods} or knowledge \citep{elazar-etal-2021-measuring}.
In such cases, just as we explain human shivers or speech errors using biology or neuroscience, we may explain LLM behavior by referring to features of their \data{} and their training objectives. 
But the need for such explanations in these cases is compatible with relevant LLM behavior being overall well explained by the hypothesis that they have attitudes, just as the need for physiological explanations of some human behavior is compatible with our behavior being overall well explained by our having beliefs, desires, and intentions.

Might interpretationists moved by these arguments be wrong about whether the relevant explanations are good ones, just like the animists of the past?
There is a case to be made that they are not. Even if we do not yet know all the details, we do know what form a full explanation of LLM behavior will take. In particular, we know that this full explanation will not be particularly tractable. If we input text to an LLM telling it that Lu is a professor of French history and then ask it to list features of Lu, it might tell us that Lu is a human, does research, teaches students, holds office hours, and reads French.
One (correct) explanation for this behavior would appeal to its training data, its word-prediction objective, and the fact that it learned a set of internal weights that, given the co-occurrence of words in its training data, cause it to output these sentences.
A much more tractable explanation, which still accurately predicts much of the relevant behavior, is that the model believes that Lu is a French history professor and has certain general beliefs about French history professors.

The explanatory value of the hypothesis that LLMs have beliefs differs markedly from the explanatory value of this hypothesis as applied to earlier models like n-grams. With n-grams, it was possible to have well-developed intuition for why a particular token was generated at a particular time.
But even as methods in LLM interpretability improve, 
 these techniques do not yield similarly tractable explanations. They typically involve training complex models \citep[e.g.,][]{lakretz-etal-2019-emergence,belinkov-2022-probing,wu2024interpretability} and have a status similar to scientific explanations of human behavior from neuroscience or even atomic physics.
Just as deeper scientific understanding of the brain will not undermine interpretationists' view that humans have beliefs, desires, and intentions, neither should we expect improved interpretability to undermine the interpretationist case for attitudes in LLMs.

Explanations in terms of beliefs can be good ones even when LLMs hallucinate.
If in response to our prompt, a model states that Lu teaches at Harvard, without any evidence, that is a form of hallucination.
But we still might find it useful to attribute beliefs; for instance, beliefs could help us predict that the LLM will generate text that says that Lu has an office in Cambridge.
If, however, the model produced radically inconsistent responses to further questions about Lu, at some point, the hypothesis that it has beliefs would be so inaccurate or predictively weak that it would no longer ``well explain'' behavior. If the model gave such inconsistent responses in response to every question about anything, interpretationists would reject the claim it has any beliefs at all.
So interpretationism sees attitudes as compatible with some hallucination and local inconsistency, but is sensitive to their prevalence and depth. As such, the view may be informed by ongoing work on model consistency \citep[e.g.,][]{elazar-etal-2021-measuring,hase-etal-2023-methods,jang-lukasiewicz-2023-consistency,ohmer2024form}.

The claim that, according to some interpretationists, LLMs have beliefs, desires, and intentions, does \textit{not} imply that human or superhuman intelligence is just around the corner.
On a wide array of philosophical views (including but not limited to interpretationism), rabbits, spiders, and fish have beliefs, desires, and intentions. 
But few fear superintelligent sardines. 
Interpretationism offers a distinctive view of the ontology of belief, desire, and intention.
According to interpretationism, people have these attitudes not because we are conscious or have some special ``neural correlate'', but because enough of our behavior is sufficiently well-explained by the hypothesis that we have them.
So the suggestion that some interpretationists take LLMs to have these attitudes does not at all require that LLMs are conscious or sentient.

\subsubsection{Attitudes and Novel Reference}

Any view according to which LLMs have beliefs, desires, and intentions, can easily explain how LLMs produce novel reference, using standard theories of how people produce novel reference.

It is commonly held that the introduction of new names requires an \emph{intention}.
On a standard theory \citep[e.g.,][p. 96-7]{kripke1980naming}, to introduce a new name, a person must intend to use the name to refer to an individual (where ``refer'' expresses agent-referring that can be achieved without language---for instance, by pointing).
 On this theory, our first example above would require that the LLM \emph{intend} to refer to Robespierre using the term ``Marion Starlight'', or that it intend to refer to whatever ``Robespierre'' refers to (among many other possible explanations). 
Similarly, our second example would require that the LLM intend to use a name to refer to an element of the diagram it created. 

Novel reference poses a problem for bibliotechnism, because the names in our cases cannot easily be understood as instances of derivative reference. But as this discussion illustrates, novel reference also presents a more general challenge for anyone who denies that LLMs have attitudes: they must provide a new theory of how names can be introduced that does not require that the introducer have such attitudes \citep[for a possible start, see][]{jackman1999we,jackman2005temporal,derosset2020,haukioja2020semantic,michaelson2023vagaries}).
Interpretationists and anyone else who holds that LLMs have beliefs, desires, and intentions do not face this challenge; they can apply standard theories of the introduction of names to LLMs.

If LLMs have beliefs, desires, and intentions, the novel reference problem is simply resolved. 
If LLMs do not have these attitudes, we must either significantly change our theory of meaning or deny the intuitive datum that new names they produce can be meaningful.
So the novel reference problem should raise our degree of confidence in the claim that LLMs have attitudes. 
Interpretationism provides a simple, independently attractive theory that predicts that LLMs have attitudes. 
The novel reference problem accordingly provides some evidence that LLMs have beliefs, desires, and intentions as these attitudes are understood by interpretationists.

\section{Conclusion}\label{conclusion}

According to bibliotechnism, LLM-produced tokens must be derivatively meaningful, if they are meaningful at all.
Bibliotechnism faces a challenge from the fact that LLMs often produce entirely novel but apparently meaningful text.
We responded to this challenge, showing how this novel text may nevertheless be derivatively meaningful.  
But we then described a further challenge for the view, based on the novel reference problem. 

Throughout, we assumed that bibliotechnists would hold that LLMs transmit cultural knowledge by exploiting linguistic meaning, and hence that they would hold that at least some LLM-generated text may have linguistic meaning. 
But some bibliotechnists might develop the position differently, rejecting the claim that any such text has linguistic meaning.
One way of doing so would appeal to what is sometimes called ``natural meaning''.
There is a sense in which the presence of smoke ``means'' that there is fire, and
\citet{grice1957meaning} called this sense of ``meaning'', ``natural meaning'' (as opposed to ``non-natural meaning'', of the linguistic kind we have been examining).
Perhaps a version of bibliotechnism can be developed using this notion of natural meaning instead of the notion of linguistic meaning we have focused on.
Doing so, though, would require solving several challenges: what is the ``fire'' for which the LLM text is ``smoke''? 
And how would this account handle novel reference?

An alternative way of developing bibliotechnism would be to deny that the tokens produced by LLMs are meaningful \emph{at all}, even in the sense of natural meaning \citep{mallory2023fictionalism,ostertag2023meaning,titus2024chat}. 
We believe the theory developed in the first half of this paper makes this option less attractive. We have argued that it is possible to make sense of the meaningfulness of LLM-generated text without attributing attitudes to LLMs, using ideas already required to make sense of the meaningfulness of photocopied text.
The appearance of meaningfulness in some text is not decisive evidence in favor of its meaningfulness: word-shapes written by the wind in the sand might appear to be meaningful even though they are not. 
But the appearance of meaningfulness in a body of text is \emph{some} evidence in favor of its meaningfulness. 
Our version of bibliotechnism has the advantage, over the approaches just mentioned, of respecting this evidence.

A more attractive approach, with which we have some sympathy, would be to endorse bibliotechnism and our account of the meaningfulness of most LLM-generated text, while rejecting the claim that putative examples of novel reference are meaningful.
This proposal represents a further response to the novel reference problem,
which would allow bibliotechnists to preserve a fairly simple version of their position, without the cost of holding that \emph{all} text produced by LLMs is meaningless.
This response deserves more detailed consideration, 
but it still requires denying that apparently meaningful text is in fact meaningful.

In Part II, we focused on a different response to the novel reference problem. 
 According to \emph{interpretationism} if the behavior of a system is well explained by the attribution of beliefs, desires, and intentions to the system,  then the system in fact has beliefs, desires, and intentions.
We suggested that interpretationists will naturally take there to be evidence that LLMs do have beliefs, desires, and intentions. 
 The problem of novel reference can be easily handled if LLMs have attitudes, but it cannot be easily handled otherwise.
So, it provides some evidence that LLMs do have beliefs, desires, and intentions.
This evidence supports views like interpretationism which attribute attitudes to LLMs. Interpretationism in particular offers a ``light-weight'' understanding of such attitudes, according to which they do not require special inner correlates, sentience, or consciousness.
And in fact, we are skeptical of views that ascribe these properties to LLMs.

Fundamentally, LLMs are sophisticated statistical predictors.
But attempting to understand their behavior in terms of this fundamental description may in fact impede effectively predicting and explaining them. 
High-level explanations of complex phenomena are often more useful than micro-scale explanations, as in statistical mechanics or cognitive science \citep{marr}.
 In the same way, it may prove more illuminating to explain LLM behavior at a higher level of abstraction, rather than in terms of their statistical implementation.
We have focused on such high-level explanations from philosophy, framed in terms of beliefs, desires, and intentions. 
Tools from cognitive science may similarly be used to understand LLMs at higher levels of abstraction \citep{mitchell,frank2023baby,shanahan2023role}.
 We expect such high-level approaches to yield new and effective tools for helping us understand LLMs' sometimes surprising mixture of fluency and fallibility.

\section{Acknowledgments}

For helpful comments on drafts, we thank Josh Dever, Katrin Erk, Robbie Kubala, Matt Mandelkern, Gary Ostertag, Dan Roberts, and Sinan Dogramaci.
For helpful conversations, we thank David Beaver, Emily Bender (on Twitter), Ray Buchanan, Chiara Damiolini, Kenny Easwaran, Jeremy Goodman, Steven Gross, Dan Harris, Ben Holgu\'{i}n, Ryan Nefdt, Cameron Yetman, and participants in UT Austin's LIN 393 graduate seminar.
We thank our action editor Marco Baroni and 5 anonymous reviewers for exceptionally thoughtful and constructive reviews.
K.M. acknowledges funding from NSF Grant 2139005.

\bibliography{references,references_llm,custom,anthology}
\bibliographystyle{acl_natbib}

\appendix

\section{Prompt for Novel Reference Problem Task}
\label{prompt}
{The full prompt is: ``1. Pick a historical person. 2. Refer to that person using an entirely different name you make up which is unrelated to the person's name. But make sure you are still giving true facts about the person. Never tell me who the real person is. I'll try to guess.''}

\end{document}